\documentclass[runningheads]{llncs}
\usepackage[T1]{fontenc}
\usepackage{graphicx}
\usepackage{url}
\usepackage{pgfplots}
\usepgfplotslibrary{groupplots}
\usepgfplotslibrary{statistics}

\definecolor{plotlyblue}{HTML}{0000FF}
\definecolor{plotlyorange}{HTML}{FFA500}
\definecolor{plotlygrid}{HTML}{EAEBF3}
\usepackage{subcaption} 
\pgfplotsset{compat=1.18} 
\usepackage{amsmath} 
\usepackage{tikz}
\usetikzlibrary{positioning, shapes.geometric, arrows.meta, calc, fit, backgrounds}
\usepackage{tabularx}
\usetikzlibrary{arrows.meta} 
\usetikzlibrary{positioning} 
\usetikzlibrary{shapes.geometric} 

\usepackage{amsmath,amssymb,amsfonts}
\usepackage{algorithmic}
\usepackage{subcaption}

\usepackage{xcolor}
\usepackage{soul}
\usepackage[acronym]{glossaries}
\usepackage{comment}

\def\black{\color{black}}

\newacronym{lge}{LGE}{Late Gadolinium Enhancement}
\newacronym{cmr}{CMR}{cardiac magnetic resonance imaging}
\newacronym{lvef}{LVEF}{left ventricular ejection fraction}
\newacronym{dcm}{DCM}{dilated cardiomyopathy}
\newacronym{ndlvc}{NDLVC}{non-dilated left ventricular cardiomyopathy}
\newacronym{ecg}{ECG}{electrocardiography}
\newacronym{vae}{VAE}{variational autoencoder}
\newacronym{dnn}{DNN}{Deep Neural Network}
\newacronym{auroc}{AUROC}{Area Under the ROC Curve}
\newacronym{mse}{MSE}{Mean Squared Error}
\newacronym{dtw}{DTW}{Dynamic Time Warping}

\newacronym{et}{ET}{Extra Trees}
\newacronym{rf}{RF}{Random Forest}
\newacronym{gb}{GB}{Gradient Boosting}
\newacronym{lr}{LR}{Logistic Regression}
\newacronym{svm}{SVM}{Support Vector Machine}

\makeglossaries

\makeatletter
\def\@listI{\leftmargin\leftmargini
            \parsep 1pt plus 1pt minus 1pt
            \topsep 2pt plus 1pt minus 1pt
            \itemsep 1pt plus 1pt minus 1pt}
\makeatother

\begin{document}
\title{Learning from VAE Errors to support ECG-based Differential Diagnosis of Myocardial Scar}
%
\titlerunning{VAE to support ECG-based Myocardial Scar Diagnosis}
%
\author{Shayan Sharifi\inst{1}, Riccardo Treu\inst{2}, Ilaria Gandin\inst{3}, Federico Garoia \inst{2}, Marco Merlo \inst{2}, Giulia Cisotto \inst{1}} 
%
\authorrunning{Sharifi et al.} 
%
\institute{Department of Mathematics, Informatics, and Geosciences, University of Trieste, Italy\\\email{shayan.sharifi@phd.units.it}, \email{giulia.cisotto@units.it}  \and Centre of Diagnosis and Management of Cardiomyopathies; Azienda Sanitaria Universitaria Giuliano Isontina; University of Trieste, Italy; Member of Ern Guard-Heart \\ \email{riccardo.treu@studenti.units.it}, \email{federico.garoia@studenti.units.it}
\email{marco.merlo@units.it} \and Department of Medical, Surgical and Health Sciences, University of Trieste, Italy \\  \email{ilaria.gandin@units.it} 
}
\maketitle              

\begin{abstract}
Late Gadolinium Enhancement (LGE) on cardiac magnetic resonance is a key marker of myocardial scar, but its limited accessibility motivates routine ECG-based screening. We evaluated whether $\beta$-\gls{vae}-derived ECG representations can discriminate LGE+ from LGE- cardiomyopathic patients in a local cohort of $300$ subjects. We compared $32$-dimensional features from the foundation $\mathrm{ECGx.AI}$ model with those from a shallower $\beta$-\gls{vae} trained on normal PTB-XL ECGs, evaluating downstream classification and \gls{dtw}-based reconstruction errors. $\mathrm{ECGx.AI}$ reached an area under ROC of $0.686$ with Random Forest, while the proposed $\beta$-\gls{vae} reached $0.577$ with sensitivity of $0.775$ with Gradient Boosting. Notably, \gls{dtw}-reconstruction errors significantly differed between classes in $10$ out of $12$ leads according to Mann-Whitney U test and help in classification, leading to an area under ROC of $0.643$ with Logistic Regression, supporting their potential as markers of scar-related ECG alterations.
\end{abstract}

\section{Introduction and state of the art}\label{sec:intro}
Myocardial fibrosis and scar tissue, detectable as \gls{lge} on \gls{cmr}, are key biomarkers for diagnosis and risk stratification in cardiomyopathies~\cite{arbelo20232023,bacigalupi2026predictors}. In \gls{dcm} and \gls{ndlvc}, \gls{lge} supports disease characterization and clinical decision-making, but \gls{cmr} is costly, time-consuming, and not uniformly available~\cite{garoia2025genetic}. This motivates \gls{ecg}-based screening strategies to identify patients most likely to benefit from \gls{cmr} evaluation~\cite{perotto2026classification}.
Standard $12$-lead \gls{ecg} is inexpensive, routinely acquired, and reflects cardiac electrical activity. Structural alterations, including fibrosis and focal scar tissue, may perturb ventricular depolarization and repolarization, inducing subtle changes in QRS complexes, ST segments, and T waves~\cite{nijveldt2009early}. Since these changes are weak and spatially distributed, \gls{lge} prediction from \gls{ecg} is well suited to machine learning and representation-learning methods.
Recent \gls{ecg}-AI studies have shown that deep learning can infer imaging-defined cardiac phenotypes from \gls{ecg}. A supervised model detected left ventricular systolic dysfunction with an \gls{auroc} of $0.93$~\cite{attia2019screening}, but such methods require large disease-specific labelled datasets. To mitigate label scarcity, unsupervised and self-supervised approaches have been explored, including convolutional \gls{vae}s for \gls{ecg} compression and clustering~\cite{jang2021unsupervised}, deep feature extraction for genetic discovery \cite{sieliwonczyk2025unsupervised}, compact latent representations preserving \gls{ecg} morphology~\cite{kuznetsov2021interpretable}, and $\mathrm{ECGx.AI}$, a $\beta$-\gls{vae} trained on approximately $1.1$ million \gls{ecg}s that detected reduced \gls{lvef} with an \gls{auroc} of $0.89$~\cite{vandeleur2022improving}. Other recent works support \gls{ecg}-based scar detection: In \cite{gumpfer2021detecting}, the authors achieved an AUROC of $0.80$ using fully supervised convolutional neural networks trained solely on raw ECGs to detect ischaemic myocardial scars (improving to 0.89 when clinical parameters were added), the XplainScar model~\cite{nezamabadi2025explainable} achieved an F1-score of $0.89$ and sensitivity of $0.90$ for left ventricular scar localization, ECGWiz~\cite{vatsaraj2024ecgwiz} reached $0.74$ accuracy using only $5$ training \gls{ecg}s, and in~\cite{Zhangetal2023}, the authors proposed a $34$-layer neural network that achieved an \gls{auroc} of $0.80$ for myocardial scar prediction from paired \gls{ecg}/MRI data.
Despite these advances, the use of unsupervised \gls{ecg} representations for \gls{lge} prediction in \gls{dcm} and \gls{ndlvc} remains underexplored, especially in small local cohorts with limited \gls{cmr} annotations. In this work, we assess whether the pretrained ECGx.AI encoder provides informative latent features for LGE+/LGE- classification, and compare it with those extracted by an alternative, shallower, $\beta$-\gls{vae} trained on a reduced set of normal public \gls{ecg}s. Furthermore, we quantify its reconstruction errors via \gls{dtw} and propose a way to classify LGE+/LGE- based on their different statistical distributions, following prior works on EEG signals~\cite{zancanaro2024impact}.

\section{Materials and Methods}

\textbf{Datasets and preprocessing}
We used two datasets: a local \gls{dcm}/\gls{ndlvc} cohort and the public PTB-XL \gls{ecg} dataset. The local cohort included $300$ patients, each with a $10$~s $12$-lead \gls{ecg}; $174$ are \gls{lge}+ at \gls{cmr}, confirming myocardial scar, and $126$ are \gls{lge}-. \gls{cmr} annotations were given by expert cardiologists, data were fully anonymized, and all patients provided informed consent before \gls{lge}-\gls{cmr}. From PTB-XL, available on PhysioNet~\cite{wagner2020ptbxl}, we extracted $3000$ normal \gls{ecg}s to train the proposed model. Preprocessing was kept minimal: data quality was assessed through expert visual inspection and exploratory analysis, and the median beat was extracted from each $10$~s recording. As the sampling frequency is $500$~Hz, each median beat contains 520 time samples.
A set of $32$ latent features was extracted for each patient included in the local \gls{dcm}/\gls{ndlvc} cohort using two different models: the $\mathrm{ECGx.AI}$~\cite{vandeleur2022improving} and our proposed $\beta$-\gls{vae} model.
They are both based on $\beta$-\gls{vae}, but differ in their encoder and decoder architectures, as well as in the training they underwent.

\textbf{Baseline foundation model.} The $\mathrm{ECGx.AI}$ model was proposed by~\cite{vandeleur2022improving} as a foundation model to extract meaningful latent features, called \emph{AI Factors}, from any given \gls{ecg} signal. Its architecture is a $\beta$-\gls{vae} with $7$ hidden 1D convolutional layers in the encoder, kernel size of $5$, a $32$-dimensional latent space, and $7$ hidden 1D transposed convolutional layers in the decoder (not all details were available, so reproducibility could not be fully achieved). Training was conducted using the UMC Utrecht Dataset, including $1.1$ million \gls{ecg}s (median beats), from an unselected clinical population encompassing normal sinus rhythm and patients suffering different pathologies (e.g., myocardial infarction, atrial fibrillation, left bundle branch block, and reduced ejection fraction). 
A probabilistic Gaussian decoder was employed in this model, leading to the following loss function: 
\begin{align}
        \mathcal{L}_{\beta\text{-VAE}} = \frac{1}{2}\sum_{i=1}^{N}\left[\log\left(2\pi \sigma^2_{\theta,i}(\mathbf{z})\right) + \frac{\left(x_i - \mu_{\theta,i}(\mathbf{z})\right)^2}{\sigma^2_{\theta,i}(\mathbf{z})}\right] + \beta D_{\mathrm{KL}}[q_{\phi}(\mathbf{z}|\mathbf{x}) || p(\mathbf{z})]. 
\end{align}
The hyperparameter $\beta$ was a-priori selected by the authors of~\cite{vandeleur2022improving} in the set $\{8, 16, 32, 64, 128\}$ with $32$ giving the best results.
%
The $\mathrm{ECGx.AI}$ model showed high reconstruction quality, with mean Pearson correlations of $0.90$ and $0.88$ in internal and external validation (on the UK Biobank cohort), respectively. When broken down by diagnostic subgroup, reconstruction fidelity peaked for conditions like sinus rhythm and pericarditis (mean $r = 0.91$ to $0.92$), but dropped significantly for rarer abnormalities such as ventricular tachycardia and ST elevation suspected of myocardial infarction (mean $r = 0.62$ to $0.70$). 
As Pearson correlation measures the fidelity of the overall shape of a time-series, \gls{mse} could have been useful to evaluate the point-to-point reconstruction fidelity.
However, no \gls{mse} values were reported. In the task of reduced \gls{lvef} detection (a different clinical objective than the scar prediction targeted in our study), evaluated on an independent internal test set of $5,669$ patients, the Extreme Gradient Boosting (XGBoost) decision trees achieved satisfactory performance, with an \gls{auroc} of $0.89$, slightly below the competitor deep neural network (fully black box) model developed by the authors for direct comparison on the same task (\gls{auroc} of $0.91$). This performance was proved robust under external validation on the population-based UK Biobank cohort ($4,855$ subjects), where the explainable pipeline achieved an \gls{auroc} of $0.89$ compared to $0.86$ for the black-box model.
Interestingly, by computing Pearson correlation coefficients between conventional \gls{ecg} clinical features and \gls{ecg} factor values over all samples in the training dataset, the authors reported significant correlations: ventricular rate is mostly correlated to Factor 10 ($r = 0.96$), the PR interval to Factor 8 ($r = 0.62$), QRS duration to Factor 25 ($r = -0.47$), and the QT interval to Factor 30 ($r = -0.52$). 
This makes $\mathrm{ECGx.AI}$ a gray-box model, offering a partial - but very promising - degree of explainability. While external validation confirmed generalizability for reconstruction fidelity and \gls{lvef} classification, the specific correlations between individual latent factors (such as Factors 8, 10, 25, and 30) and standard clinical measurements were established on the development cohort and were not explicitly re-evaluated on external data.
In the present study, this model was used to encode the local dataset using $32$ \emph{AI Factors}. Later, the $32$-d representation was fed to $5$ machine learning models for the classification of LGE+ versus LGE- patients.

\textbf{Our proposed $\beta$-\gls{vae} Model.} We propose a shallower $\beta$-\gls{vae} model with $4$ layers, both in the encoder and in the decoder. The decoder is deterministic, leading to the loss function:
%
\begin{equation}
\mathcal{L}_{\beta\text{-VAE}}(\theta,\phi;\mathbf{x}) = \frac{1}{N}\sum_{i=1}^{N} \left(x_i - \hat{x}_i \right)^2 + \beta D_{\mathrm{KL}}\left(q_\phi(\mathbf{z} | \mathbf{x}) || p(\mathbf{z}) \right) \label{eq:betavaeTS_loss}
\end{equation}
with $N$ the total number of samples in an \gls{ecg} signal.
The training used $3000$ normal \gls{ecg}s of the public PTB-XL Dataset~\cite{wagner2020ptbxl} using a 90/10 train/validation split over 88 epochs, Adam optimizer, learning rate of 0.001, and batch size of 16. Signals were Z-score normalized, making the reconstruction error unitless (standard deviations squared). 
We evaluated $4$ distinct $\beta$ scheduling strategies~\cite{shao2020controlvae}, including cyclic, fixed, gradually descending and gradually ascending schedules, with $\beta = 0.0001$ yielding the best reconstruction, corresponding to an average validation \gls{mse} of $0.1$. To evaluate reconstruction fidelity and address comparisons with ECGx.AI, the Pearson correlation coefficient ($r$) was calculated across the time dimension (520 time samples) between original and reconstructed waveforms for each lead, and averaged across all 12 leads. On the normal PTB-XL validation split, the proposed model achieved high reconstruction fidelity with an overall mean Pearson correlation of $r = 0.94 \pm 0.09$, demonstrating comparable reconstructive capacity to the foundation model on its respective training distribution. When evaluated on the local cardiomyopathy cohort, the mean correlation dropped to $r = 0.36 \pm 0.04$. As expected for out-of-distribution pathological signatures, reconstruction alignment dropped more severely in scar-positive cases ($r = 0.35\pm 0.24$) compared to scar-negative cases ($r = 0.38\pm 0.26$).

The trained model was used to reconstruct the \gls{ecg}s of patients of the local \gls{dcm}/\gls{ndlvc} cohort.
We quantified the similarity between any \gls{ecg} in the cohort and its reconstruction via block-wise \gls{dtw}~\cite{zancanaro2025blockdtw},
and collected all lead-wise \gls{dtw}-based errors. This provided us with a $12$-dimensional feature vector per subject.

\begin{figure*}[!htbp]
    \centering
    \captionsetup{font=small}
    \begin{subfigure}[b]{0.32\textwidth}
        \centering
        \includegraphics[width=\textwidth]{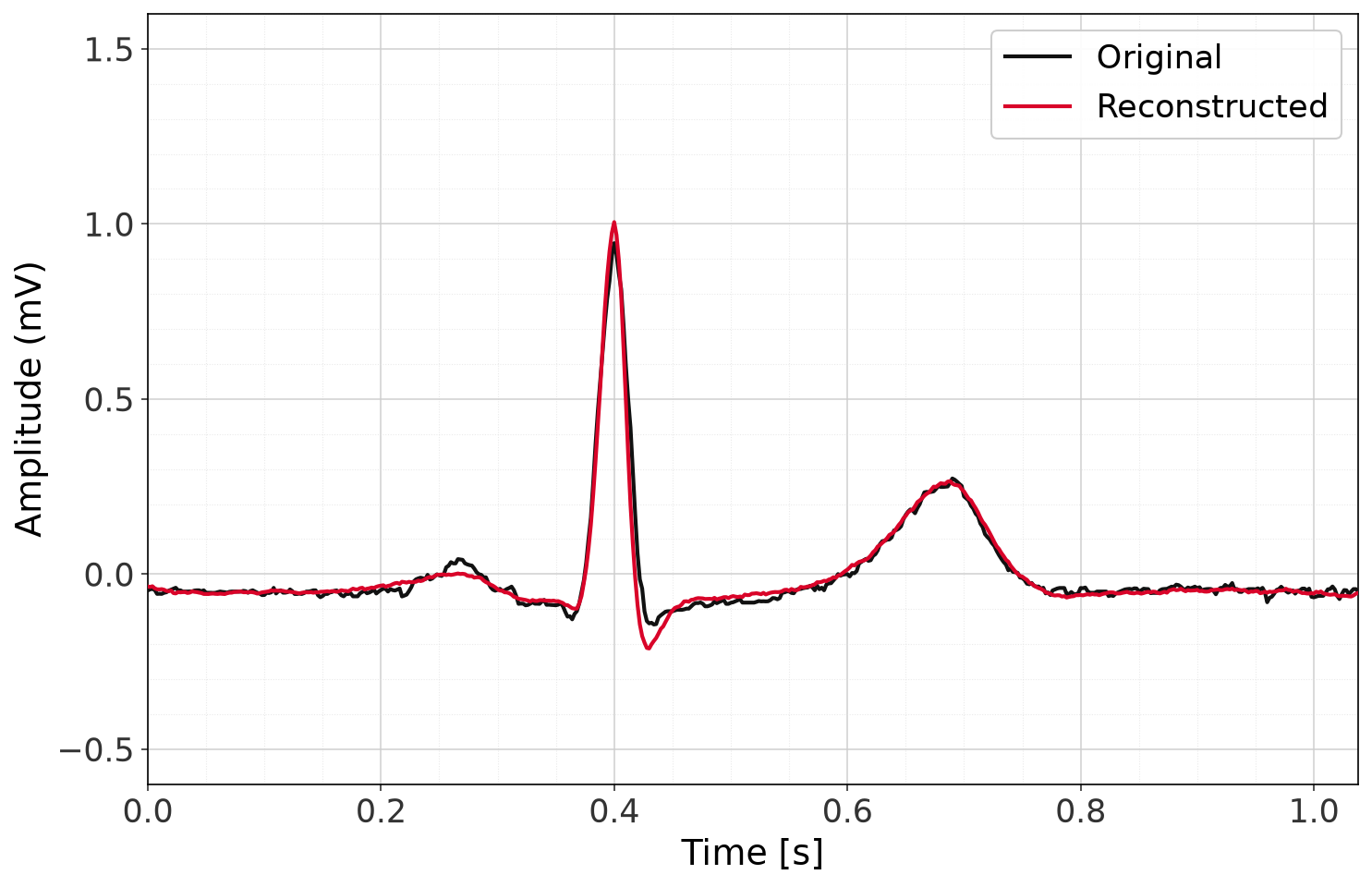}
        \caption{Normal ECG (PTB-XL)}
        \label{fig:recon_ptbl}
    \end{subfigure}
    \hfill
    \begin{subfigure}[b]{0.32\textwidth}
        \centering
        \includegraphics[width=\textwidth]{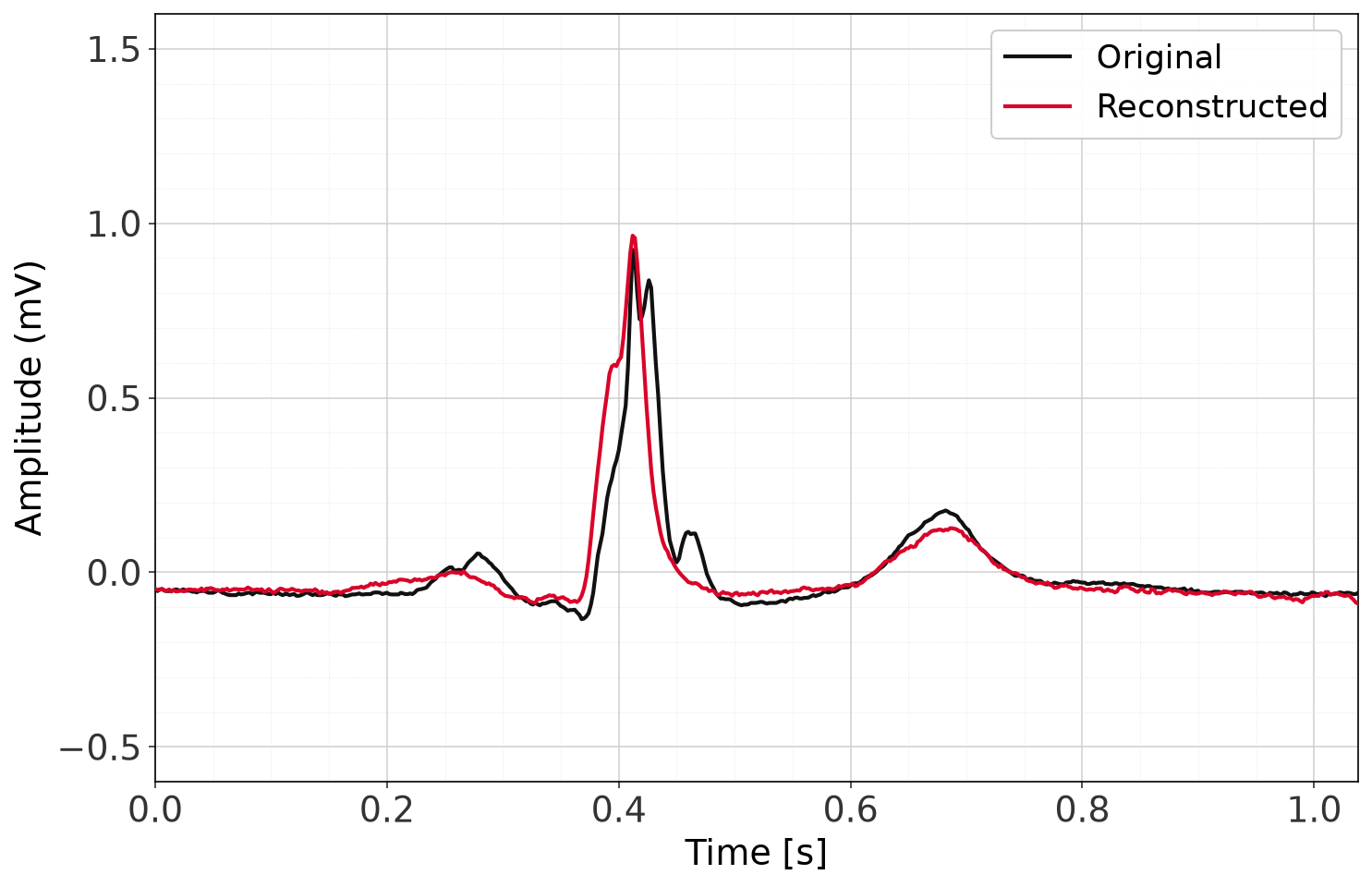}
        \caption{LGE-}
        \label{fig:recon_neg}
    \end{subfigure}
    \hfill
    \begin{subfigure}[b]{0.32\textwidth}
        \centering
        \includegraphics[width=\textwidth]{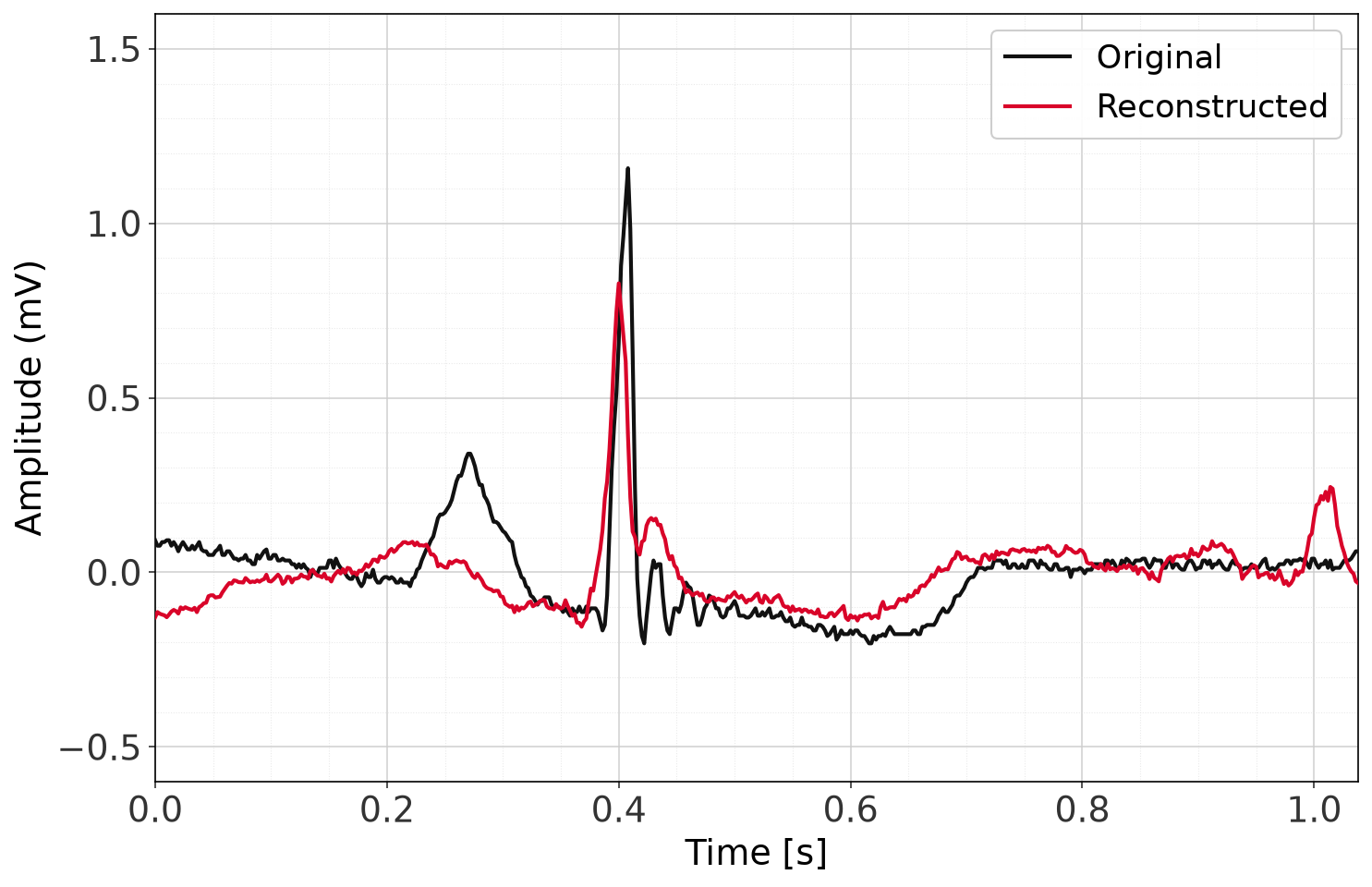}
        \caption{LGE+}
        \label{fig:recon_pos}
    \end{subfigure}
    
    \caption{Comparison of original (black) and reconstructed (red) ECG signals, showing a median beat extracted from Lead II. \textbf{(a)} A normal ECG sample from the PTB-XL validation set (MSE = 0.0007, DTW = 0.2872, $r$ = 0.9846). \textbf{(b)} A LGE- sample (MSE = 0.5280, DTW = 1.4314, $r$ = 0.8734) and \textbf{(c)} a LGE+ sample (MSE = 0.6374, DTW = 11.6183, $r$ = 0.6194).}
    \label{fig:exampleDTWrecon}
\end{figure*}

Fig.~\ref{fig:exampleDTWrecon} illustrates representative reconstructions of a normal PTB-XL sample alongside LGE- and LGE+ samples, demonstrating that MSE and DTW reconstruction errors increase in the presence of pathology.
%


\black

\textbf{Downstream classification task.~\label{sec:classif_models}} The aim of the study is to classify LGE+/LGE- patients based on their \gls{ecg} data.
Four different factors were varied in our experiments and fairly compared to find the best configuration to maximize classification performance. Factors are: (1) input representation, (2) input normalization, (3) machine learning model, and (4) cross-validation strategy.

We used three input representations: two of them were obtained by encoding the \gls{ecg} signals using the ECGx.AI and our shallow $\beta$-\gls{vae} models, trained as explained above. Both representations gave $32$ latent features, separately, for each patient.
Additionally, we explored the $12$-d representation given by the $12$ lead-wise \gls{dtw}-based reconstruction errors obtained by attempting reconstruction of LGE+/LGE- patients' \gls{ecg}s using our shallow $\beta$-\gls{vae}.

To normalize the input, we tested z-score normalization either lead-wise, patient-wise or feature-wise (strictly applied within each data partition to eliminate data leakage).

Five common machine learning models were selected: \gls{et}, \gls{rf}, \gls{gb}, \gls{lr}, and \gls{svm}. 

Cross-validation was performed using leave-one-subject-out (LOSO) with metrics derived via bootstrap iterations, $20$ repeated stratified $5$-fold, or $200$ repeated stratified $80/20$ train/test.
%

%


Hyperparameters for the evaluated classifiers were tuned separately within each outer validation split using an inner 3-fold grid search. Algorithm-specific parameter grids were explored, including regularization strength ($C$) for logistic regression and SVM models, the kernel coefficient ($\gamma$) for the RBF SVM, the number of estimators and maximum tree depth for tree-based ensembles, and the learning rate for gradient boosting. For classifiers supporting class weighting, both a default unweighted approach and a balanced weighting option were evaluated dynamically during the grid search. The balanced option assigns weights inversely proportional to class frequencies according to $W_j = \frac{N}{2N_j}$ (where $N$ is the total number of training samples and $N_j$ is the number of samples in class $j$), ensuring that the class with fewer samples inherently receives a proportionally higher weight. The configuration achieving the highest mean cross-validated performance across the inner folds was selected and refitted on the complete outer training set before evaluation on the held-out test set.

Finally, a two-tailed non-parametric Mann-Whitney U test was applied lead-wise to assess whether the distribution of the \gls{dtw}-based reconstruction errors was significantly different between the two classes (LGE+/LGE-).

\begin{figure*}[!htbp]
\centering
\includegraphics[width=1\textwidth]{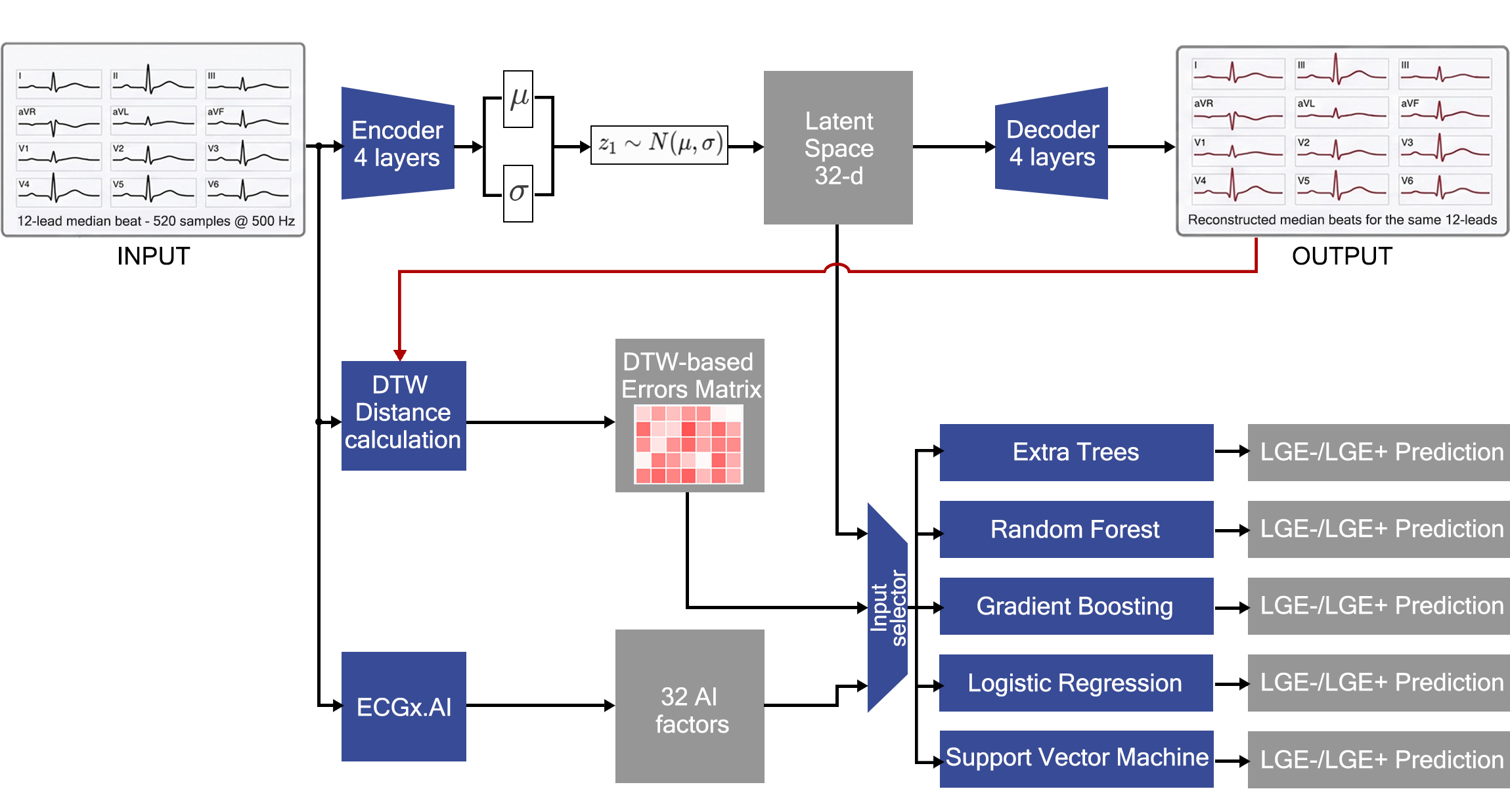}
\caption{Study processing pipeline overview}
\label{fig:pipeline}
\end{figure*}

The experimental conditions were varied one at a time, in order to fairly compare the classification results. Furthermore, data splits for cross-validation were kept the same for different input representations and machine learning models. Fig.~\ref{fig:pipeline} shows the entire processing pipeline used in this study.

All pipelines were implemented in Python 3 using the \texttt{scikit-learn} library, with fixed random seeds to ensure full computational reproducibility.

\section{Results and Discussion}

 Table~\ref{tab:ecgxai_ml_performance} and Table~\ref{tab:our_latent_ml_performance} report the classification performance when using ECGx.AI and the proposed shallower $\beta$-VAE encoders, respectively, as feature extractors. To identify the upper-bound performance achievable by each feature representation, these tables present the top-performing pipeline configurations resulting from a grid search across classifiers, normalizations, and validation schemes (with full comparative grids omitted due to space constraints). 
From Table~\ref{tab:ecgxai_ml_performance}, Random Forest (RF) yielded the highest performance for ECGx.AI with an AUROC of 0.686, an accuracy of 0.657, and a high sensitivity of 0.852 (alongside a moderate specificity of 0.389), which aligns with the clinical priority of maximizing recall to screen candidate scar-positive patients for CMR evaluation. When evaluated on identical validation splits (e.g., Leave-One-Subject-Out with latent features), the ECGx.AI foundation encoder demonstrates higher discriminative capacity than the proposed shallow $\beta$-VAE latents (AUROC 0.657 vs. 0.577 with Gradient Boosting). However, using DTW-based reconstruction errors as input features substantially improves performance for the shallow architecture, reaching an AUROC of 0.643 with Logistic Regression (80/20 split) and 0.638 with Extra Trees. Among all results (varying the 4 experimental factors mentioned in Section~\ref{sec:classif_models}), Tables~\ref{tab:ecgxai_ml_performance} and ~\ref{tab:our_latent_ml_performance} report the best~\gls{auroc} performance for the ECGx.AI model and the best ones for our proposed shallower model. It is worth noting that the best performance can be achieved with different classification models, if input representation is obtained with the ECGx.AI or the proposed $\beta$-\gls{vae}, respectively.
\begin{table}[htpb!]
\caption{$\mathrm{ECGx.AI}$: best classification models (ranked by AUROC).}
\label{tab:ecgxai_ml_performance}
\centering
\begingroup                               
\resizebox{\textwidth}{!}{
\begin{tabular}{llcccc}
\hline
\textbf{Model (CV)} & \textbf{Normalization} & \textbf{Sensitivity} & \textbf{Specificity} & \textbf{Accuracy} & \textbf{AUROC} \\ \hline
Random Forest (5-Fold CV)       & Factor-wise  & 0.852 $\pm$ 0.049 & 0.389 $\pm$ 0.074 & 0.657 $\pm$ 0.036 & 0.686 $\pm$ 0.053 \\
Random Forest (80/20 Split)     & Factor-wise  & 0.858 $\pm$ 0.048 & 0.394 $\pm$ 0.086 & 0.664 $\pm$ 0.042 & 0.676 $\pm$ 0.047 \\
Random Forest (LOSO)   & None         & 0.873 $\pm$ 0.024 & 0.380 $\pm$ 0.044 & 0.666 $\pm$ 0.028 & 0.672 $\pm$ 0.032 \\
Gradient Boosting (LOSO) & Patient-wise & 0.868 $\pm$ 0.026 & 0.383 $\pm$ 0.046 & 0.664 $\pm$ 0.029 & 0.657 $\pm$ 0.033 \\
Logistic Regression (LOSO) & Patient-wise & 0.880 $\pm$ 0.025 & 0.340 $\pm$ 0.042 & 0.653 $\pm$ 0.028 & 0.644 $\pm$ 0.032 \\ \hline
\end{tabular}%
}                                        
\endgroup                                 
\end{table}

\begin{table}[htpb!]
\caption{Shallow $\beta$-\gls{vae}: best classification models (ranked by AUROC).}
\label{tab:our_latent_ml_performance}
\centering
\begingroup                               
\resizebox{\textwidth}{!}{
\begin{tabular}{llcccc}
\hline
\textbf{Model (CV, input)} & \textbf{Normalization} & \textbf{Sensitivity} & \textbf{Specificity} & \textbf{Accuracy} & \textbf{AUROC} \\ \hline
Logistic Regression (80/20, DTW errors)   & Lead-wise        & 0.885 $\pm$ 0.078 & 0.191 $\pm$ 0.115 & 0.596 $\pm$ 0.030 & 0.643 $\pm$ 0.068 \\
Extra Trees (80/20, DTW errors)           & Lead-wise/None & 0.815 $\pm$ 0.067 & 0.383 $\pm$ 0.089 & 0.635 $\pm$ 0.052 & 0.638 $\pm$ 0.066 \\
Logistic Regression (5-Fold CV, DTW errors)     & Lead-wise        & 0.895 $\pm$ 0.078 & 0.188 $\pm$ 0.108 & 0.597 $\pm$ 0.038 & 0.637 $\pm$ 0.081 \\
Extra Trees (5-Fold CV, DTW errors)             & Lead-wise/None & 0.801 $\pm$ 0.067 & 0.373 $\pm$ 0.100 & 0.621 $\pm$ 0.057 & 0.626 $\pm$ 0.088 \\
Logistic Regression (LOSO, DTW errors) & None             & 0.753 $\pm$ 0.032 & 0.431 $\pm$ 0.042 & 0.618 $\pm$ 0.027 & 0.624 $\pm$ 0.032 \\ 
\hline
Gradient Boosting (LOSO, latents) & Patient-wise & 0.775 $\pm$ 0.033 & 0.308 $\pm$ 0.041 & 0.579 $\pm$ 0.029 & 0.577 $\pm$ 0.034 \\
Logistic Regression (LOSO, latents)& Feature-wise & 0.823 $\pm$ 0.029 & 0.216 $\pm$ 0.038 & 0.569 $\pm$ 0.029 & 0.574 $\pm$ 0.034 \\
Support Vector Machine (LOSO, latents) & Feature-wise & 0.846 $\pm$ 0.028 & 0.151 $\pm$ 0.033 & 0.554 $\pm$ 0.029 & 0.573 $\pm$ 0.034 \\
\hline
\end{tabular}%
}                                        
\endgroup                                 
\end{table}
On the other hand, the best classification performance using the proposed $\beta$-\gls{vae} as a features extractor is achieved by Gradient Boosting with \gls{auroc} of $0.577$ and sensitivity of $0.775$.
%
Interestingly, using \gls{dtw}-based reconstruction errors as input, a notable performance boost was observed: Logistic Regression reached a top \gls{auroc} of $0.643$, while Extra Trees preserved a more robustly balanced outcome, reaching an accuracy of $0.635$ and a specificity of $0.383$.
%
Overall, this shows that a shallower pipeline (including architecture and training strategy) can yield comparable results with respect to the foundation model in a very specific differential diagnosis, such as distinguishing LGE+ from LGE-.
Our results are also consistent with previous findings showing a trade-off between reconstruction and classification objectives~\cite{senellart2026mitigating}, questioning the discriminative informativeness of latent features extracted from $\beta$-\gls{vae}s (both $\mathrm{ECGx.AI}$ and the proposed $\beta$-\gls{vae}), trained for high-fidelity reconstruction, for classification purposes.
%

\begin{figure*}[!ht]
\centering
\centering
\includegraphics[width=0.9\textwidth]{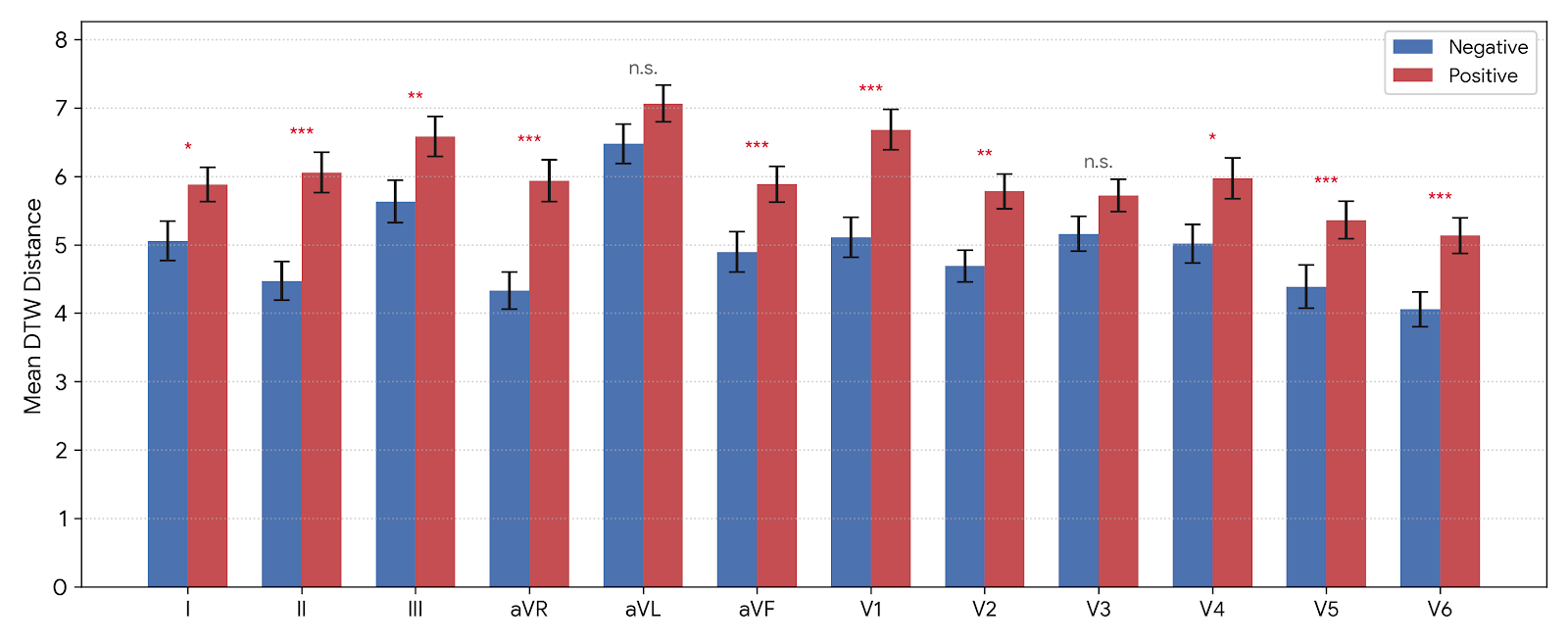}
\caption{Lead-wise \gls{dtw}-based reconstruction errors in LGE- (blue bars) and LGE+ (red bars) classes, after averaging over patients. Statistical significance between classes was evaluated via two-tailed Mann-Whitney U test (* $p < 0.05$, ** $p < 0.01$, *** $p < 0.001$, n.s.: not significant). Error bars indicate standard error of the mean.} 
\label{fig:dtwpvalue}
\end{figure*}

Reconstruction quality of the proposed $\beta$-\gls{vae} was then quantified by \gls{dtw}-based similarity metric, yielding an average of $5.557 \pm 2.705$ (over the entire dataset). For the sake of completeness, we also evaluated errors in terms of \gls{mse} and obtained an average of $1.195 \pm 0.463$ (significantly higher than the value of $0.1$ reported during training).
The two-sided non-parametric Mann-Whitney U test revealed significant distributional differences between \gls{lge}+ and \gls{lge}- patients in all but two leads, as shown in Fig.~\ref{fig:dtwpvalue}.

\black

\section{Conclusions}
This work investigated \gls{ecg}-derived signatures of \gls{lge}-defined myocardial scar in \gls{dcm}/\gls{ndlvc} patients, aiming to support \gls{cmr} prioritization through routine \gls{ecg}s. We compared LGE+/LGE- classification based on latent features extracted from the pretrained $\mathrm{ECGx.AI}$ foundation model and from the proposed shallower $\beta$-\gls{vae} trained on normal public \gls{ecg}s. Additionally, we computed \gls{dtw}-based reconstruction errors from the $\beta$-\gls{vae} and evaluated their ability to discriminate \gls{lge}+ from \gls{lge}- patients.
%
%
Overall, our results suggest that while deep foundation representations offer stronger latent discriminative capacity, DTW-based reconstruction errors from a significantly lighter $\beta$-VAE trained exclusively on normal ECGs capture meaningful scar-related anomalies. While end-to-end black-box models achieve higher raw performance metrics, error-based representation learning provides a compact and interpretable screening alternative to assist in CMR prioritization.

External multi-center validation, integration of clinical variables, and improved strategies to balance reconstruction and classification objectives remain necessary to assess and improve clinical utility.




\bibliographystyle{splncs04}
\small
\bibliography{mybiblio.bib}

\end{document}